\documentclass[letterpaper]{article} 
\usepackage{aaai25}  
\usepackage{times}  
\usepackage{helvet}  
\usepackage{courier}  
\usepackage[hyphens]{url}  
\usepackage{graphicx} 
\usepackage{natbib}  
\usepackage{caption} 
\usepackage{algorithm}
\usepackage{algorithmic}

\usepackage{appendix}
\usepackage{amsmath} 
\usepackage{amsfonts,amssymb} 
\usepackage{booktabs} 
\usepackage{multirow}
\usepackage{makecell}
\usepackage{circledsteps}
\usepackage{bbding}

\usepackage{newfloat}
\usepackage{listings}
\DeclareCaptionStyle{ruled}{labelfont=normalfont,labelsep=colon,strut=off} 
\floatstyle{ruled}
\newfloat{listing}{tb}{lst}{}
\floatname{listing}{Listing}
\title{Dual Conditioned Motion Diffusion for Pose-Based Video Anomaly Detection}
\author{
    Hongsong Wang\textsuperscript{\rm 1,2},
    Andi Xu\textsuperscript{\rm 3},
    Pinle Ding\textsuperscript{\rm 3},
    Jie Gui\textsuperscript{\rm 3,\rm 4,\rm 5} \thanks{Corresponding Author}
}
\affiliations{
    \textsuperscript{\rm 1}School of Computer Science and Engineering, Southeast University, Nanjing 210096, China \\
     \textsuperscript{\rm 2}Key Laboratory of New Generation Artificial Intelligence Technology and Its Interdisciplinary Applications (Southeast University), Ministry of Education, China \\
    \textsuperscript{\rm 3}School of Cyber Science and Engineering, Southeast University, Nanjing 210096, China\\
    \textsuperscript{\rm 4}
    Engineering Research Center of Blockchain Application, Supervision And Management (Southeast University), Ministry of Education, China \\
    \textsuperscript{\rm 5}
    Purple Mountain Laboratories, Nanjing 210000, China \\
     \{hongsongwang, andixu, pinleding, guijie\}@seu.edu.cn \\
}

\begin{document}

\maketitle

\begin{abstract}
Video Anomaly Detection (VAD) is essential for computer vision research. Existing VAD methods utilize either reconstruction-based or prediction-based frameworks. The former excels at detecting irregular patterns or structures, whereas the latter is capable of spotting abnormal deviations or trends. We address pose-based video anomaly detection and introduce a novel framework called Dual Conditioned Motion Diffusion (DCMD), which enjoys the advantages of both approaches. The DCMD integrates conditioned motion and conditioned embedding to comprehensively utilize the pose characteristics and latent semantics of observed movements, respectively. In the reverse diffusion process, a motion transformer is proposed to capture potential correlations from multi-layered characteristics within the spectrum space of human motion. To enhance the discriminability between normal and abnormal instances, we design a novel United Association Discrepancy (UAD) regularization that primarily relies on a Gaussian kernel-based time association and a self-attention-based global association. Finally, a mask completion strategy is introduced during the inference stage of the reverse diffusion process to enhance the utilization of conditioned motion for the prediction branch of anomaly detection. Extensive experiments on four datasets demonstrate that our method dramatically outperforms state-of-the-art methods and exhibits superior generalization performance. 
\end{abstract}

\begin{links}
    \link{Code}{https://github.com/guijiejie/DCMD-main}
\end{links}

\section{Introduction}
\begin{figure}[t]
  \centering
  \includegraphics[width=\linewidth]{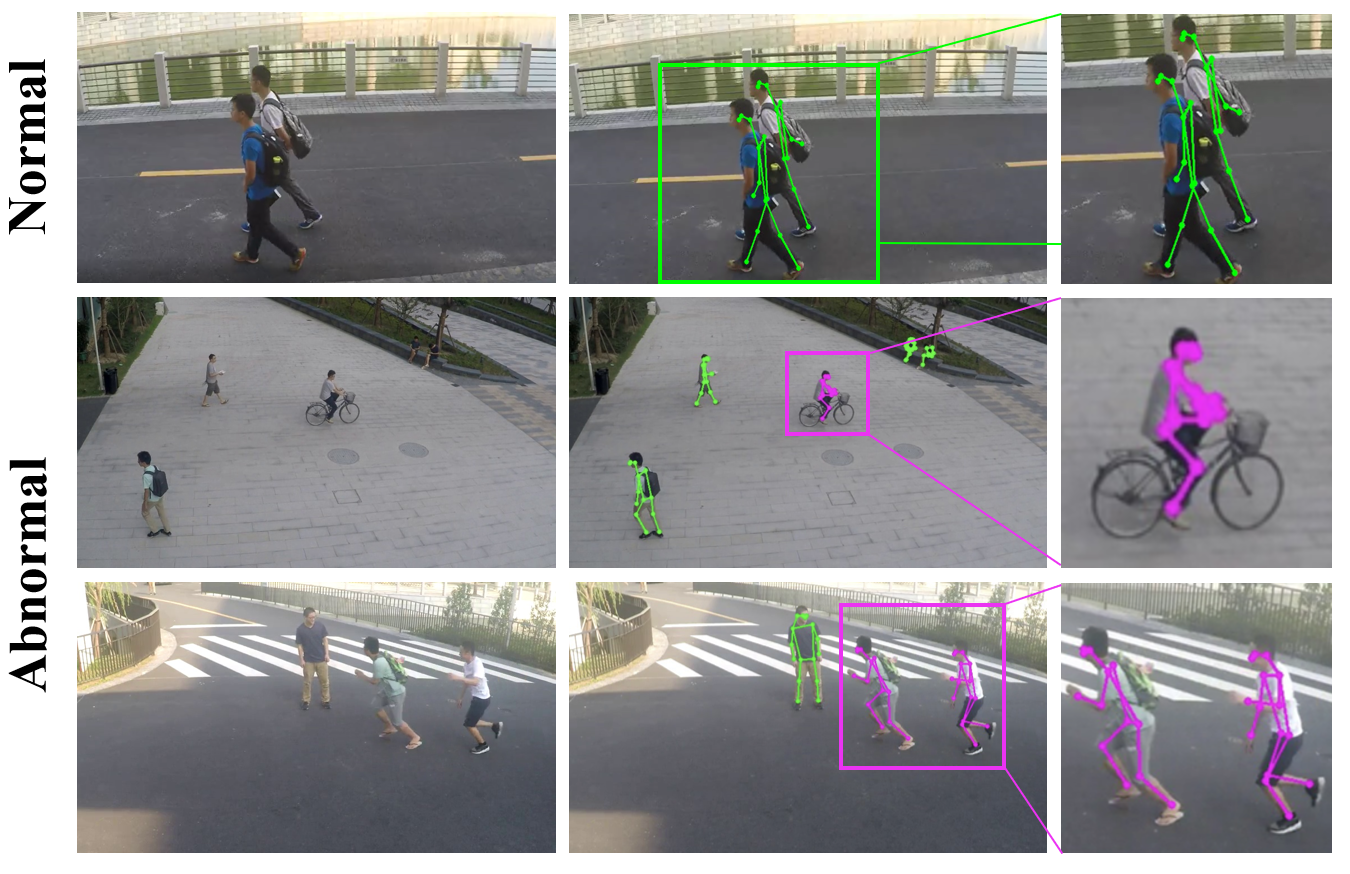}
  \caption{Illustration of video anomaly detection with normal instances (green) and abnormal instances (red).}
  \label{fig:introduction}
\end{figure}

Video Anomaly Detection (VAD) is an essential topic in computer vision and security applications. Different from human action understanding~\shortcite{wang2018beyond,chen2023occluded,wang2024region,weng2024usdrl}, VAD enables prompt identification of abnormal occurrences, such as human actions, accidents, and illnesses. Anomalies are generally characterized as uncommon, unexpected, or unusual phenomena that exhibit significant deviations from normality. Conversely, normality is defined as what is expected and regularly encountered. 
Detecting video anomalies can be challenging as these events occur infrequently and belong to various categories. Labeling data is costly and time-consuming, making collecting all possible abnormal samples impractical for fully supervised learning approaches. Consequently, anomaly detection problems are typically treated as One Class Classification (OCC) \shortcite{Zaheer2020OldIG}. Only normal data is utilized during model training, and data that deviates significantly from the normal pattern is identified as an anomaly during testing \shortcite{Cai2021AppearanceMotionMC}.

There has been a growing trend toward detecting abnormal events or behaviors by analyzing human poses extracted from video frames \shortcite{Flaborea2023MultimodalMC}, as illustrated in Figure \ref{fig:introduction}. Utilizing skeletons to depict human motions in videos is a highly effective approach to protecting privacy and circumventing the limitations of appearance-based attributes. Besides privacy protection, skeletal features are compact, well-structured, and highly descriptive of human motion \shortcite{Morais2019LearningRI}.  

Existing VAD works are mainly categorized into reconstruction-based and prediction-based methods. Reconstruction-based methods \shortcite{Li2021SpatialTemporalCA, Chang2020ClusteringDD} first compress the input data into a low-dimensional representation and then recover the original data from this representation. By comparing the reconstruction error between the original data and the recovered data, anomalies such as irregular patterns or structures in the data can be detected. 
Prediction-based methods \shortcite{Liu2017FutureFP} focus on predicting future frames or events based on history data, and anomalies are indicated when future data deviates from the predicted trends. These methods are effective in modeling and uncovering the temporal connections between successive frames. Nevertheless, they may be prone to noise \shortcite{Tang2020IntegratingPA}.

Recently, diffusion models have shown great success in generating high-quality samples \cite{Ho2020DenoisingDP}. One of the outstanding capabilities of diffusion models is the flexibility to handle a wide range of input conditions that can encompass different types of data, enabling customized generation tasks \cite{Ho2022ClassifierFreeDG}. Due to the inherent multimodality of human motion and the diversity of both normal and abnormal patterns, diffusion models are naturally well-suited for modeling human motion and detecting anomalies. 
However, simply applying diffusion models for VAD suffers inherent limitations of reconstruction-based or prediction-based approaches. 

To address the above issues, we present a unified framework that seamlessly combines the advantages of both reconstruction-based and prediction-based approaches for pose-based VAD. The reconstruction is based on an auto-encoder architecture, whereas the prediction utilizes a diffusion model. 
We propose a novel Dual Conditioned Motion Diffusion (DCMD) that seamlessly integrates conditioned motion and conditioned embedding, enabling the exploitation of both pose characteristics and latent semantics of observed movements. During the reverse diffusion process, we introduce a motion transformer specifically designed to extract potential correlations from multi-layered spectrum features of human motion. In addition to reconstruction and prediction losses, we devise a United Association Discrepancy (UAD) regularization that leverages Gaussian kernel-based time association and self-attention-based global association. Furthermore, during the inference stage, we employ a mask completion strategy to bolster the utilization of conditioned motion for prediction-based anomaly detection. Experiments conducted on popular human-related anomaly detection datasets demonstrate the superior performance of our proposed method compared to state-of-the-art approaches. 

Our main contributions are summarised below:
\begin{itemize}
    \item We introduce a novel framework that seamlessly integrates reconstruction-based and prediction-based methods for video anomaly detection, leveraging the strengths of both approaches.

    \item We propose a Dual Conditioned Motion Diffusion (DCMD), which incorporates both conditioned motion and conditioned embedding in a diffusion-based model.

    \item We propose a motion transformer for anomaly detection with a novel regularization that uses both time association and global association to improve the discriminability between normal and abnormal instances.

    \item We present a mask completion method during the denoising process of diffusion, enabling more effective utilization of observed motions while predicting future motions for anomaly detection.
\end{itemize}

\section{Related Work}

We summarize previous work on Video Anomaly Detection (VAD) from three aspects: reconstruction-based, prediction-based, and pose-based. 

\noindent{\textbf{Reconstruction-Based Methods:}}
Reconstruction-based methods comprise two components: an encoder and a decoder. The encoder compresses the input frame into a low-dimensional feature while the decoder reconstructs the output from this compressed representation. Reconstruction error is a criterion for distinguishing between normal and abnormal events.
Luo et al. \shortcite{Luo2019VideoAD} introduce a novel deep neural network architecture that leverages sparse coding, incorporates a temporal coherence term to preserve similarity between similar frames, and employs a stacked recurrent neural network to optimize sparse coefficients for achieving real-time anomaly detection. 
Li et al. \shortcite{Li2021SpatialTemporalCA} present a two-stream network designed to capture the visual and motion characteristics of typical events in videos. 
To encode the scene, objects as well as motion information, Chang et al. \shortcite{Chang2020ClusteringDD} propose a novel deep k-mean clustered convolutional self-encoder architecture.
There are also some methods utilizing architectures other than convolutional neural networks. 
Gong et al. \shortcite{Gong2019MemorizingNT} introduce a memory-enhanced autoencoder that can distinguish new test samples by memorizing prototypical normal data elements during training. 
Doshi et al. \shortcite{Doshi2022RethinkingVA} propose a dual-stage approach combining deep learning with a kNN-based RNN to overcome forgetting in end-to-end models, enabling efficient continual learning.
Zhong et al. \shortcite{zhong2022cascade,zhong2022reverse} design a cascade reconstruction model and spatio-temporal autoencoder. 

\begin{figure*}[t]
  \centering
  \includegraphics[width=\linewidth]{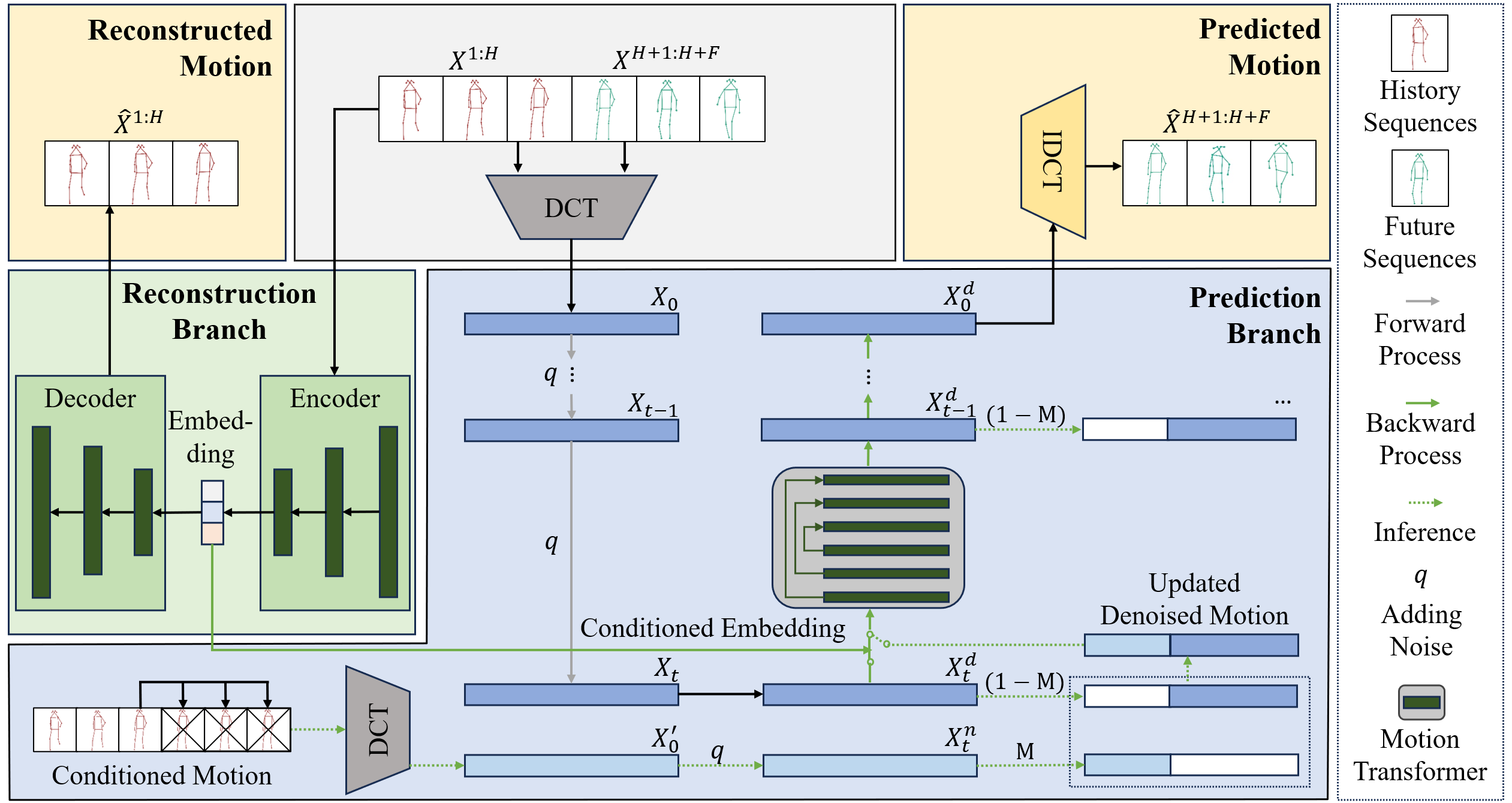}
  \caption{Overall architecture of the proposed method. The $H+F$ skeletal motion sequences are split into history motion sequences (red skeletal motion sequences) and future motion sequences (green skeletal motion sequences). The key point of our model is the dual conditioned motion diffusion, i.e., the hidden representation of the observed sequence obtained by the encoder and the complete sequence of observed sequences with added noise connected to the predicted future motion sequence.
}
  \label{fig:architecture}
\end{figure*}

\noindent{\textbf{Prediction-Based Methods:}}
In prediction-based methods, the models are often employed to forecast the next frame by employing a sequence of preceding frames as inputs. Prediction-based methods are more effective in analyzing spatio-temporal patterns between frames than reconstruction-based methods. 
Liu et al. \shortcite{Liu2017FutureFP} pioneer prediction-based VAD, using U-Net to predict future frames and incorporating motion constraints for consistent optical flow. 
Zhou et al. \shortcite{Zhou2020AttentionDrivenLF} adopt a similar network architecture and propose an attention-driven loss algorithm to address the challenge of imbalanced foreground targets and static backgrounds in anomaly detection videos. 
Doshi et al. \shortcite{doshi2021online} present an online anomaly detection method that consists of a feature extraction module and a statistical decision-making module. 
To consider both spatio-temporal characteristics, Lee et al. \shortcite{Lee2018StanST} utilize a spatio-temporal generator that synthesizes an inter-frame with bidirectional ConvLSTM. 
Different from the memory module in \cite{Gong2019MemorizingNT}, 
Park et al. \shortcite{Park2020LearningMN} introduce a novel memory module to capture normal prototype features and incorporate a wider range of patterns. 
There are also two-stream approaches that separately learn spatio-temporal normality patterns. 
Chang et al. \shortcite{Chang2022VideoAD} design a two-stream model consisting of a spatial autocoder and a temporal autocoder based on deep K-mean clustering. 
Hao et al. \shortcite{Hao2022SpatiotemporalCN} leverage a 3D CNN-based encoder and a 2D CNN-based decoder to improve the consistency of generated results in the spatio-temporal domain. 
Cai et al. \shortcite{Cai2021AppearanceMotionMC} utilize the prior knowledge of appearance and motion signals to explicitly capture their correspondence in the high-level feature space. 
A bidirectional spatio-temporal feature learning framework is also proposed~\shortcite{zhong2022bidirectional}.
Although the aforementioned methods make numerous attempts to acquire spatio-temporal representations and achieve improved results, prediction-based methods are still limited in certain aspects. A potential solution to this challenge lies in combining the strengths of both prediction-based and reconstruction-based methods.

\noindent{\textbf{Pose-Based Methods:}}
Instead of using RGB video, pose-based methods utilize low-dimensional semantically skeleton data which simulates the dynamics of human joints over time. 
Morais et al. \shortcite{Morais2019LearningRI} capture the overall dynamics of the body in motion and the spatial relationships of the skeleton. 
Markovitz et al. \shortcite{Markovitz2019GraphEP} use embedded pose graphs and a Dirichlet process mixture for pose-based anomaly detection and introduce a coarse-grained setting aiming to detect abnormal variations of an action.
Luo et al. \shortcite{Luo2020NormalGS} propose a novel technique using a spatio-temporal Graph Convolutional Network (GCN).  
Yu et al. \shortcite{yu2023regularity} introduce a motion prior regularity learner to enhance dynamic representation.
Jain et al. \shortcite{Jain2021PoseCVAEAH} recently introduce an innovative strategy of the Conditional Variational Auto-Encoder (CVAE) framework, which employs a hybrid training strategy that combines self-supervised and unsupervised learning. 
Flaborea et al. \shortcite{Flaborea2023ContractingSK} utilize a graph convolutional network to represent human skeletal motion and learn to encode skeletal kinematics onto a minimum volume of the potential hypersphere. Unlike the aforementioned approaches, we investigate the roles of conditions in the diffusion model for VAD, along with the examination of training and inference strategies. 

\section{Method}

Consider $X=\left [x^1,\cdots,x^{H+F}\right ] \in \mathbb{R}^{\left(H+F\right)\times J\times C}$ as a series of $H+F$ continuous motion sequences belonging to a participant. Here, $x^t\in\mathbb{R}^{J\times C}$ refers to the joint coordinates in frame $t$, $J$ represents the number of joints, and $C$ denotes the pose's dimension. We divide $X$ into two parts: history motion sequence $X^{1:H}$ and future motion sequence $X^{H+1:H+F}$. The architecture of the proposed model is illustrated in Figure \ref{fig:architecture}. The model comprises two branches, namely the prediction branch and the reconstruction branch. The former involves a diffusion process and a reverse process to make predictions, whereas the latter employs an encoder-decoder structure to reconstruct the history motion.

\begin{figure}[t]
  \centering
  \includegraphics[width=\linewidth]{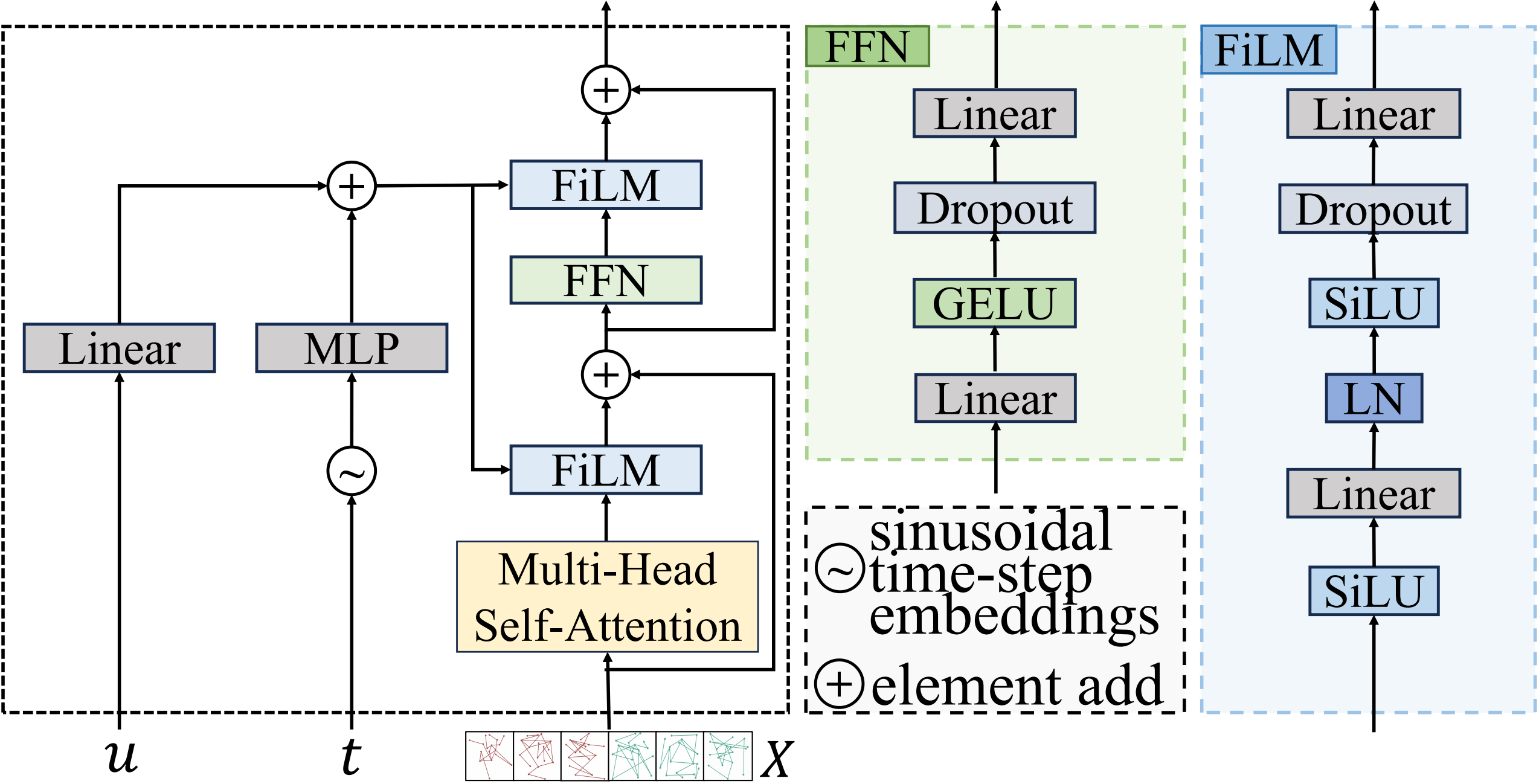}
  \caption{The architecture of the denoising network Motion Transformer. The green box on the right describes the details of the FFN module, and the blue box describes the details of the FiLM module.}
  \label{fig:anotrans}
\end{figure}

\subsection{Dual Conditioned Motion Diffusion}
We construct a new Dual Conditioned Motion Diffusion (DCMD) for human motion synthesis. The key idea is to consider conditioned embedding and conditioned motion as the dual conditions in the reverse process to gradually predict the future sequence from noisy variable distribution. We assume that the network of noise prediction is parameterized with $\theta$. 
Let $\mathrm {E}$ denote the reconstruction branch encoder. 
In the reverse process of the proposed, the recovered human motion is 
\begin{equation}
    X_{t-1} = \frac{1}{\sqrt{\alpha_{t}}} \left ( X_{t}-\frac{1-\alpha_{t}}{\sqrt{1-\bar{\alpha}_t}}\epsilon_{\theta}\left ( X_{t},t,u \right ) \right ) + \sigma_{t} z ,
\end{equation}
where $u$ denotes motion encoding, which contains content information of the history motion sequence, i.e., $u=\mathrm {E}(X^{1:H})$, we name $u$ as \textbf{conditioned embedding}. 

Instead of recovering human motions from raw random noise, we add noise to the history motion sequence, concatenate it with the predicted future motion sequence, and subsequently send the whole sequence into the reverse process. We name the noised history sequence as \textbf{conditioned motion}. Details are shown in the Inference Section.

The denoising network in the reverse process is denoted as \textbf{motion transformer}, which is characterized by allowing for acquiring potential correlations from multi-layered characteristics. $L$ motion transformer blocks incorporate skip connection stacking. Each block contains two linear-based FiLM modules inspired by \cite{Perez2017FiLMVR}. The FiLM modules are modulated by the diffusion time-step $t$ embedding and conditioned embedding to establish temporal relationships. Its overall structure is shown in Figure~\ref{fig:anotrans}. 
Assuming the input motion sequence is $X$, the overall equation can be formalized as follows:
\begin{equation}
\begin{aligned}
    Z&=\text {FiLM}\left(\text {Attention}\left(X\right)+\text{TE}\left(t\right)+u\right) + X,\\
    Y&=\text {FiLM}\left(\text {FFN}\left(Z\right)+\text{TE}\left(t\right)+u\right) + Z,
\end{aligned}
\end{equation}
where $Y\in {\mathbb{R}^{(H+F)\times D} }$ denotes an output with channel $D$. $Z\in {\mathbb{R}^{(H+F)\times D} }$ is the hidden representation. $\mathrm {TE}(\cdot )$ denotes sinusoidal time-step embedding, and $\mathrm {Attention}(\cdot )$ denotes multi-head self-attention which is calculated by
\begin{equation}
\begin{aligned}
    \text {Attention}\left(X\right)=\text {Softmax}(\frac{{{Q_m}{K_m^T}}}{\sqrt{2D} } ){V_m} ,
\end{aligned}
\end{equation}
where ${Q_m=XW_Q}, {K_m=XW_K}, {V_m=XW_V} \in \mathbb{R}^{(H+F)\times{\frac{D}{h}}}$ is the query, key, and value of the $m$-th head self-attention, respectively. $\mathrm {Softmax}(\cdot )$ normalizes the attention graph along the last dimension.

Finally, motion transformer concatenates the outputs of multi-heads $\left \{ {Y}_m \in \mathbb{R} ^{(H+F)\times\frac{D}{h} } \right \}_{1\le m\le h}$ to obtain the final result $Y$.

\begin{figure}[t]
  \centering
  \includegraphics[width=\linewidth]{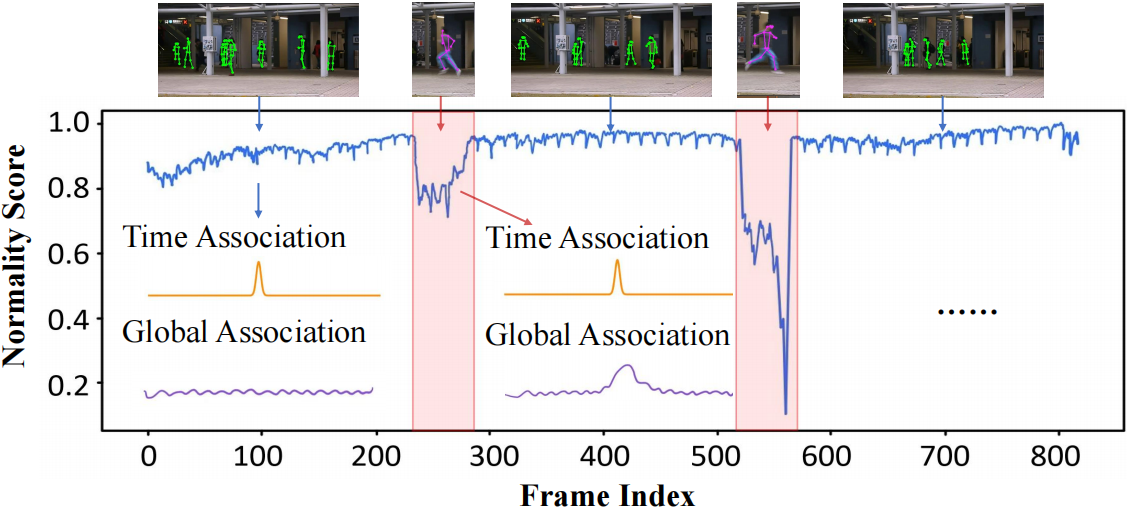}
  \caption{Graphical illustration of time association and global association. The blue curve is the normality score for a test segment of the CHUK dataset, and the red area indicates the time period in which the abnormal event occurred. Yellow curves indicate time association, and purple curves indicate global association.}
  \label{fig:TA_GA}
\end{figure}

\begin{algorithm}[t]
	\renewcommand{\algorithmicrequire}{\textbf{Require:}}
	\renewcommand{\algorithmicensure}{\textbf{Output:}}
	\caption{Training procedure of the proposed framework.}
	\label{alg1}
	\begin{algorithmic}[1]
		\REQUIRE noising steps $T$, maximum iterations $I_{max}$.
            \renewcommand{\algorithmicrequire}{\textbf{Input:}}
            \REQUIRE motion sequence $X \in \mathbb{R}^{(H+F)\times 2J}$.
            \ENSURE the noise prediction network $\epsilon_{\theta}$.
            \FOR{$I \gets 0$ to $I_{max}$}
                \STATE Divide $X$ into history motion sequence $X^{1:H}$ and future motion sequence $X^{H+1:H+F}$.
		      \STATE Encode the history motion sequence as $u$.
		      \STATE Decode $u$ as reconstruction motion sequence $\hat{X}^{1:H}$.
                \STATE Calculate the reconstruction loss $\mathcal{L}_{rec}$.
		      \STATE Perform DCT transformation on $X$ yields ${X_0}$.
                \STATE Sample the time steps $t=\mathrm {Uniform}\left ( \left \{ 1,2,\dots ,T \right \}  \right )$.
		      \STATE Add $t$-step noise to $X_0$ using pre-defined variance parameters $\alpha_t$ and noise $\epsilon \in \mathcal{N}(0,I)$ yields ${X_t}$.
                \STATE Calculate the prediction loss $\mathcal{L}_{pred}$ in Eq.~(\ref{eq:pred_loss}).
                \STATE Calculate the reconstruction loss $\mathcal{L}_{rec}$ in Eq.~(\ref{eq:rec_loss}).
		      \STATE Calculate the additional loss UAD in Eq.~(\ref{eq:loss_uad}).
		      \STATE Calculate the total loss $\mathcal{L}_{total}$ in Eq.~(\ref{eq:loss_total}).
                \STATE Update parameters of the network
                
                $\theta =\theta -\nabla_{\theta}\mathcal{L}_{total}$.
            \ENDFOR
	\end{algorithmic}  
\end{algorithm}

\subsection{Training with UAD}
Inspired by \cite{Mao2019LearningTD}, during the training stage, we perform a Discrete Cosine Transform (DCT) operation on the complete motion sequence $X$ to acquire the spectrum, 
\begin{equation}
    {X_0} = \text{DCT}\left( X \right),
\end{equation}
where $\mathrm{DCT(\cdot)}$ denotes the DCT operation, and $X_0$ is the DCT coefficients.

In the forward process $q$, we can compute the noisy DCT spectrum $X_t$ at a given time step $t$ by employing the reparameterization trick.
\begin{equation}
    X_{t} = \sqrt{\bar{\alpha}_t}X_{0}+\sqrt{1-\bar{\alpha}_t}\epsilon ,
\end{equation}
where $\bar{\alpha}_{t}= {\textstyle \prod_{i=1}^{t}}\alpha_{i} $, $\alpha_{i} \in \left [ 0,1 \right ]$ are pre-defined variance parameters, and $\epsilon \sim \mathcal{N}(0,I) $.

In the reverse process, we optimize the denoising model parameter $\theta$ using the noise prediction loss function,
\begin{equation}
    \mathcal{L}_{pred} = smoot{h_{{L_1}}}(\mathbb{E} _{\epsilon,t}\left [ \left \| \epsilon - \epsilon_{\theta}\left ( X_{t},t,u \right )\right \|^{2} \right ]) ,
    \label{eq:pred_loss}
\end{equation}
where $smoot{h_{{L_1}}}( \cdot )$ denotes the Smooth L1 Loss.

Considering both the temporal and spatial dimensions of the input motion sequence, we use the Space-Time-Separable Autoencoder (STSAE) \cite{Sofianos2021SpaceTimeSeparableGC, Flaborea2023MultimodalMC}, which relies on a U-Net-like architecture \cite{Wyatt2022AnoDDPMAD} to shrink the skeletal motion network step-wise and expand the spatial dimensions of the input motion sequence. The encoder $\mathrm {E}$ encodes the history motion sequence as $u=\mathrm {E}(X^{1:H})$, while the decoder $\mathrm {D}$ reconstructs them into $\hat{X}^{1:H}$ by using the reconstruction loss,
\begin{equation}
    \mathcal{L}_{rec} = \left \| \hat{X}^{1:H}-X^{1:H} \right \|_{2}^{2} .
    \label{eq:rec_loss}
\end{equation}

In order to enhance the distinction between normal and abnormal frames, we leverage the \textbf{United Association Discrepancy (UAD)} as an additional loss function. The primary components of the UAD loss consist of \textbf{time association} and \textbf{global association}, both aimed at serving the subsequent United Association Discrepancy. The two components are illustrated in Figure \ref{fig:TA_GA}.

Time association is only considered from the dimension of time distance. Frames close to each other should have a stronger association, while those farther apart should have a weaker one. This pattern follows the normal distribution curve, resulting in a single-peaked shape. To establish time association, we employ a learnable Gaussian kernel \cite{Xu2021AnomalyTT} to determine relative time distance beforehand. Besides, we incorporate a learnable scale parameter to the Gaussian kernel, which enables us to concentrate on the neighboring regions for time association and accommodate various time-series patterns, including anomalies of varying durations. The formulation is as follows:
\begin{equation}
    \mathcal{T} = \text {Rescale}\left ( \left [ \frac{1}{\sqrt{2\pi}\sigma_{i}}\mathrm{exp}\left ( -\frac{\left | j-i \right | ^2 }{2\sigma_{i}^2} \right ) \right ] \right ),
\end{equation}
where $\sigma \in{{\mathbb{R}}^{(H+F)\times{h}}}$ denotes the learnable scale parameter for $h$ heads and $i,j\in \left \{ 1,\cdots,(H+F) \right \}$. To transform the association weights into a discrete distribution, we utilize $\mathrm {Rescale}(\cdot )$ by dividing the row sum.

Global association is considered from the numerical dimension. Frames with abnormal behavior should have a stronger association with the frames close to each other in the time dimension and a weaker association with the frames farther away, showing a single-peaked pattern. Similarly, frames that behave normally have a strong association in the time dimension with frames that are either far or close in distance. Global association learns associations from the original motion sequence and subsequently adaptively identifies the most efficient associations. Global association is part of self-attention, which is calculated as:
\begin{equation}
    \mathcal{G} = \text {Softmax}(\frac{{{Q_m}{K_m^T}}}{\sqrt{2D} } ) .
\end{equation}

\begin{algorithm}[t]
	\renewcommand{\algorithmicrequire}{\textbf{Require:}}
	\renewcommand{\algorithmicensure}{\textbf{Output:}}
	\caption{Inference procedure of the proposed framework.}
	\label{alg2}
	\begin{algorithmic}[1]
		\REQUIRE noising steps $T$, the mask of the observation $\mathrm {M}$, the trained noise prediction network $\epsilon_{\theta}$.
            \renewcommand{\algorithmicrequire}{\textbf{Input:}}
            \REQUIRE observed motion sequence $X^{1:H} \in \mathbb{R}^{H\times 2J}$.
            \ENSURE completed motion sequence ${X} \in \mathbb{R}^{(H+F)\times 2J}$.
            \STATE Encode the $X^{1:H}$ sequence as $u$.
            \STATE Sample random noise $X_{T}^d \in \mathcal{N}(0,I)$.
            \STATE Pad the observed motion sequence $X^{'}\in \mathbb{R}^{(H+F)\times 2J}$.
            \STATE Perform DCT transformation on $X^{'}$ yields ${X_0^{'}}$.
            \FOR{$t \gets T-1$ to $0$}   
                \STATE Select Gaussian noise $z\in \mathcal{N}(0,I)$ if $t>0$, else set $z=0$.
                \STATE Add $t$-step noise to $X_0^{'}$ yields noised history sequence $X_{t}^{n}$.
                \STATE Select denoised prediction sequence $X_t^d$.
                \STATE Perform mask completion for predicted motions
                $\mathrm {DCT}\left [ \mathrm {M}\odot \mathrm {iDCT}\left (X_{t}^{n}\right )+\left ( 1-\mathrm {M} \right ) \odot \mathrm {iDCT}\left (X_{t}^{d}\right )\right ]$. 
            \ENDFOR
            \STATE Perform iDCT transformation on $X_0$.
            \STATE Select the following F frames as predicted motion sequence $\hat{X}^{H+1:H+F}$.
            \STATE Decode $u$ as reconstructed motion sequence $\hat{X}^{1:H}$.
            \STATE Concatenate the reconstructed motion and the predicted motion to obtain the completed motion $X$.
	\end{algorithmic}  
\end{algorithm}

\begin{table*}[ht]
  \centering
    \begin{tabular}{llcccc}
    \toprule
          &       & HR-STC & HR-Avenue & HR-UBnormal & UBnormal \\
    \midrule
    MPED-RNN \cite{Morais2019LearningRI} & \multicolumn{1}{c}{\textit{CVPR’19}} & 75.4  & 86.3  & 61.2  & 60.0 \\
    GEPC \cite{Markovitz2019GraphEP} & \multicolumn{1}{c}{\textit{CVPR’20}} & 74.8  & 58.1  & 55.2  & 53.4 \\
    PoseCVAE \cite{Jain2021PoseCVAEAH} & \multicolumn{1}{c}{\textit{ICPR’21}} & 75.7  & 87.8  & --     & -- \\
    MoCoDAD \cite{Flaborea2023MultimodalMC} & \multicolumn{1}{c}{\textit{ICCV’23}} & 77.6  & 89.0  & 68.4  & 68.3 \\
    COSKAD \cite{Flaborea2023ContractingSK} & \multicolumn{1}{c}{\textit{PR’24}} & 77.1  & 87.8  & 65.5  & 65.0 \\
    TrajREC \cite{stergiou2024holistic} & \multicolumn{1}{c}{\textit{WACV’24}} & 77.9 &  89.4 & 68.2 & 68.0 \\
    \midrule
    \multicolumn{2}{l}{Reconstruction Only} &  77.1 & 87.4 & 66.4 & 66.2 \\
    \multicolumn{2}{l}{Prediction Only} &  77.5 & 88.7& 67.3 & 67.2 \\
    \multicolumn{2}{l}{\textbf{DCMD (Ours)}} & \textbf{78.6} & \textbf{90.0} &  \textbf{69.0}  & \textbf{69.0} \\
    \bottomrule
    \end{tabular}
  \caption{Comparison of our proposed method with the state-of-the-art methods based on pose data for the AUC score (\%).}
  \label{tab:result}
\end{table*}

Abnormal frames exhibit a more significant similarity between time association and global association, resulting in a smaller KL value. Conversely, normal frames display lower similarity between these associations, yielding a larger KL value. As a result, we define the United Association Discrepancy as a symmetric KL divergence between time association and global association \cite{Xu2021AnomalyTT}. UAD is computed by averaging over multiple layers, thereby consolidating the associations of a range of feature layers into a more informative and robust metric,
\begin{equation}
\begin{aligned}
    &\text{UAD}\left( {\mathcal{T},\mathcal{G};X} \right) = \\
    &{\left[ {\frac{1}{L}\mathop \sum \limits_{l = 1}^N (\text{KL}(\mathcal{T}_{t,:}||\mathcal{G}_{t,:}) + \text{KL}\left( {\mathcal{G}_{t,:}||\mathcal{T}_{t,:}} \right))} \right]_{t = 1, \ldots ,H + F}} ,
\end{aligned}
\label{eq:loss_uad}
\end{equation}
where the KL divergence is computed between the two discrete distributions that correspond to each row of $\mathcal{T}$ and $\mathcal{G}$.
We can see that abnormal frames have a smaller UAD than normal frames, making it an inherently distinguishable factor. 
Therefore, the total loss becomes:
\begin{equation}
    \mathcal{L}_{total}=\mathcal{L}_{rec}+\mathcal{L}_{pred}-\lambda \times \left \| \text{UAD}\left (\mathcal{T},\mathcal{G};X \right ) \right \|_{1} ,
    \label{eq:loss_total}
\end{equation}
where $\lambda$ is the contribution of additional loss.

The primary objective of our study is to minimize the total loss. In cases where the parameter $\lambda > 0$, we aim to optimize the additional loss by enhancing the UAD. However, directly maximizing the UAD is problematic because the abnormal frames are few and can be easily overlooked. Meanwhile, the scale parameter of the Gaussian kernel is significantly reduced, rendering the time association irrelevant. To better control united learning, we utilize the minimax strategy for UAD, which employs a specially designed stopping gradient mechanism to constrain time association and global association for more distinguishable united association discrepancy. Specifically, during the UAD minimization phase, we aim to make $\mathcal{T}$ converge towards $\mathcal{G}$, which is learned from the original motion sequence. $\mathcal{G}$ stops gradient backpropagation, and this convergence process helps $\mathcal{T}$ adjust different temporal patterns. In the UAD maximization phase, we focus on optimizing $\mathcal{G}$ by stopping the gradient backpropagation of $\mathcal{T}$ in order to increase the UAD, thereby compelling $\mathcal{G}$ to place more emphasis on non-adjacent frames. The training process of the proposed method is described in Algorithm \ref{alg1}.

\subsection{Inference with Mask Completion} 

During the inference stage, the poses of all participants in all frames of the input motion sequences window $\mathcal{W}$ are first extracted to obtain the set $\mathcal{A}$. The reconstruction branch generates $m$ history motion sequence by utilizing the encoder in conjunction with the symmetric decoder. Meanwhile, the prediction branch generates $m$ future motion sequence via mask completion. Specifically, the last frame of the observation sequence is first repeated until it matches the length of the complete sequence. The padded sequence is then transformed by DCT to get compact history information, denoted as $X_{0}^{'}$. The noise is added to $X_{0}^{'}$ to derive the noise spectrum of the observation at time step $t$, which is denoted as $X_{t}^{n}$.

Unlike previous methods, we aggregate the noised history sequence and denoised prediction sequence because it can be observed that the noise observation spectrum, $X_{t}^n$, and the denoised prediction spectrum, $X_{t}^d$, are approximately distributed similarly. The iDCT operation is employed to transform the noised and denoised spectrum into the time domain. Then, a mask mechanism is employed to combine the two spectrums,
\begin{equation}
    X_{t} = \mathrm {DCT}\left [ \mathrm {M}\odot \mathrm {iDCT}\left (X_{t}^{n}\right )+\left ( 1-\mathrm {M} \right ) \odot \mathrm {iDCT}\left (X_{t}^{d}\right )\right ] ,
\end{equation}
where $\mathrm{M}$ is the mask with the first $H$ values being 1 and the other values being 0, and $\odot$ denotes the Hadamard product.

Subsequently, this updated denoised motion and the conditioned embedding are treated as dual conditions and fed into the denoising network. Therefore, we can complete the predicted motion for each reverse diffusion step. The inference process is described in Algorithm \ref{alg2}.

\section{Experiments}

\subsection{Implementation Details}
We conduct experiments on four popular benchmarks: Human-related ShanghaiTech Campus (\textbf{HR-STC}), Human-related CUHK Avenue (\textbf{HR-Avenue}), \textbf{HR-UBnormal}, and \textbf{UBnormal}. To avoid influences caused by incorrect skeleton detection in video frames, we followed the setting~\cite{Morais2019LearningRI} that removes segments where skeletons cannot be detected using pose estimation algorithms. The Area Under Curve (AUC) of the Receiver Operating Characteristic (ROC) curve is used as the evaluation metric. 

Human motion is represented using a 17-joint skeleton. For extracting motion sequences, we employ a window size of 7 frames, where the first 3 frames comprise the historical motion sequences, and the subsequent 4 frames represent the future motion sequences. We train the network end-to-end using the Adam optimizer with a learning rate of $1e-4$ that is decayed every 36 epochs. The diffusion process employs cosine variance scheduling with ${\beta}_1=1e-4$, ${\beta}_T=2e-2$, and $T=10$. We set $\lambda=0.01$. The hidden sizes of the encoder for the reconstruction branch are (512, 256), and the dimension of the hidden embedding is 256. The noise prediction network consisted of 6 layers of motion transformer blocks, where the number of heads is 8, and the hidden dimension is 512. The experiments are conducted on an NVIDIA GeForce RTX 4090 GPU. The batch size is set to 4096 for HR-STC and 1024 for HR-Avenue. The training process lasts approximately 6 hours.

\begin{table}[ht]
    \centering
    \belowrulesep=0pt
    \aboverulesep=0pt
	\begin{tabular}{c|ccc|c}
	\toprule
            & UAD & CE &  MC  &  HR-Avenue (AUC) \\
	\midrule
		Ours & \Checkmark & \Checkmark  & \Checkmark   & \textbf{90.0} \\
        \midrule
		\textcircled{1} &   & \Checkmark  & \Checkmark   & 89.1 \\
		\textcircled{2} & \Checkmark &   & \Checkmark   & 88.7 \\
		\textcircled{3} & \Checkmark & \Checkmark &    & 89.1 \\
		\textcircled{4} &   &   & \Checkmark   & 87.2 \\
		\textcircled{5} &   & \Checkmark  &    & 87.8 \\
		\textcircled{6} & \Checkmark  &   &    & 87.0 \\
		\textcircled{7} &  &   &  & 86.7 \\
        \midrule
		\textcircled{8} &\multicolumn{2}{l}{w/o DCT}&& 89.2 \\
	\bottomrule
	\end{tabular}
 \caption{Ablation Studies on the HR-Avenue dataset (\%).}
 \label{tab:ablation}
\end{table}

\subsection{Comparison with the State-of-the-Arts}
Our approach is compared with recent state-of-the-art methods such as PoseCVAE \cite{Jain2021PoseCVAEAH}, COSKAD \cite{Flaborea2023ContractingSK}, and MoCoDAD \cite{Flaborea2023MultimodalMC}. Specifically, PoseCVAE \cite{Jain2021PoseCVAEAH} and COSKAD \cite{Flaborea2023ContractingSK} are reconstruction-based approaches.  
Specifically, MoCoDAD \cite{Flaborea2023MultimodalMC} applies the diffusion model to generate future motions based on past motions. 

Table \ref{tab:result} shows the comparison of our approach with state-of-the-art methods on the HR-STC, the HR-Avenue, the HR-UBnormal, and the UBnormal. Our approach consistently beats the other state-of-the-art methods on the four benchmarks. Our approach outperforms the recent diffusion-based method MoCoDAD \cite{Flaborea2023MultimodalMC} by 1.0\%, 1.0\%, 0.6\%, and 0.7\% in AUC scores on the HR-STC, HR-Avenue, HR-UBnormal, and UBnormal datasets, respectively. Furthermore, our method significantly surpasses the reconstruction-based methods, specifically outperforming COSKAD \cite{Flaborea2023ContractingSK} by 3.5\% and 4.0\% in AUC scores on the HR-UBnormal and UBnormal datasets, respectively. The results clearly demonstrate the advantages of our approach for video anomaly detection. 

As our approach performs video anomaly detection by combining reconstructed and predicted errors of both history and future frames, respectively, we also present the results of anomaly detection solely based on either the reconstruction or the prediction branch. Our results surpass those of any single branch, demonstrating the complementarity of reconstruction and prediction for video anomaly detection. 

\subsection{Ablation Studies}
In order to more comprehensively demonstrate the efficacy of our proposed framework, we undertake ablation studies, examining factors such as DCT, United Association Discrepancy (UAD), Conditioned Embedding (CE), and Mask Completion (MC). We report the AUC scores for our model and its variants on HR-Avenue \cite{Morais2019LearningRI} in Table \ref{tab:ablation}.
Specifically, \textcircled{7} indicates the baseline of the diffusion-based model that generates motion only from pure noise. 

After removing the DCT, the result on the HR-Avenue dataset decreases 0.8\%, it is interpreted that incorporating DCT enhances the accuracy of the motion predictions, thus improving the prediction branch for VAD. 
By removing the UAD module, we find that the AUC score decreases by 0.9\%, which shows that UAD enhances the distinction between normal and abnormal frames. 
When we remove the CE module, the whole framework amounts to prediction only, and we notice a 1.3\% decrease in the AUC score, demonstrating the necessity of combining reconstruction and prediction methods. 
Removing the MC module results in a decrease in the AUC score of 0.9\%, suggesting that the mask completion operation can effectively utilize the observed frames to generate more controlled predictions. 
In addition, we further demonstrate the necessity of each module by removing both UAD and CE, UAD and MC, and CE and MC, the AUC scores decrease by 2.8\%, 2.2\%, and 3.0\%. 
All results show the necessity of each module, and by combining each module, our model obtained the highest AUC score.

\begin{table}[t]
  \centering
    \begin{tabular}{cccc}
    \toprule
    Parameter & Value & HR-STC & HR-Avenue \\
    \midrule
    \multirow{3}*{\makecell{History}} 
          & 2     & 77.0   & 86.7   \\
          & \textbf{3}     & \textbf{78.6} & \textbf{89.9} \\
          & 4     & 77.8 & 88.0 \\
          & 5     & 77.7 & 87.1 \\
    \midrule
    \multirow{3}*{\makecell{Future}} 
          & 3     & 77.9 & 89.9 \\
          & \textbf{4}     & \textbf{78.1} & \textbf{90.0} \\
          & 5     & 77.8 & 88.5 \\
    \midrule
    \multirow{4}*{$\lambda$} & 0  & 77.6 & 89.1 \\
          & \textbf{0.01} &\textbf{77.9} & \textbf{90.0} \\
          & 0.02  & 77.8 & 88.6 \\
          & 0.05  & 75.6 & 87.0 \\
    \bottomrule
    \end{tabular}
  \caption{Parameter analysis of the proposed method.}
  \label{tab:addlabel}
\end{table}

\subsection{Parameter Analysis}
We explore the impacts of the length of history and future motion sequence, and the value of the additional loss parameter on the HR-STC and HR-Avenue datasets, respectively. Our best method is highlighted in bold in Table \ref{tab:addlabel}. Since we employ the sliding window technique, the length of each of our windows should not be too large to enhance the accuracy of our computations. Specifically, for both datasets, we select a history motion sequence length of 3 and a future motion sequence length of 4. Besides, we present a reasonable selection of different values of additional loss parameters, and the most reasonable value is 0.01. When the value increases, the result decreases significantly.

\section{Conclusion}
In this work, we propose a Dual Conditioned Motion Diffusion (DCMD) for Video Anomaly Detection (VAD), which combines the advantages of both reconstruction-based and prediction-based methods. The DCMD incorporates conditioned motion and conditioned embedding in the diffusion-based network of future motion prediction. During training, a United Association Discrepancy (UAD) regularization is introduced, and during inference, a mask completion method is employed. Our approach consistently achieves state-of-the-art performance on four VAD datasets. Detailed ablated experiments demonstrate the effectiveness of different components of the DCMD. Further experimental analysis demonstrates that the reconstruction and prediction branches are very complementary in VAD. Parameter analysis also suggests that the proposed framework is not sensitive to parameters. We hope this research can bring a new perspective on VAD by combining the advantages of both reconstruction-based and prediction-based methods.

\appendix
\section{Acknowledgments}
This work was supported by National Science Foundation of China (62172090, 62302093, 52441503), Jiangsu Province Natural Science Fund (BK20230833), Start-up Research Fund of Southeast University (RF1028623097), and Big Data Computing Center of Southeast University.

\bibliography{aaai25}

\end{document}